\documentclass[conference]{IEEEtran}
\IEEEoverridecommandlockouts
\usepackage{cite}
\usepackage{amsmath,amssymb,amsfonts}
\usepackage{algorithmic}
\usepackage{graphicx}
\usepackage{textcomp}
\usepackage{xcolor}

\usepackage{enumerate} 
\usepackage{subcaption}
\usepackage{graphicx}
\usepackage{amsmath}
\usepackage{booktabs}
\usepackage{algorithm}
\usepackage{algorithmic}
\usepackage{enumitem}

\def\BibTeX{{\rm B\kern-.05em{\sc i\kern-.025em b}\kern-.08em
    T\kern-.1667em\lower.7ex\hbox{E}\kern-.125emX}}
\begin{document}

\title{A Continuous Information Gain Measure to Find the Most Discriminatory Problems for AI Benchmarking}

\author{\IEEEauthorblockN{Matthew Stephenson}
\IEEEauthorblockA{\textit{Department of Data Science and}\\ \textit{Knowledge Engineering}\\
\textit{Maastricht University}\\
Maastricht, the Netherlands \\
matthew.stephenson@maastrichtuniversity.nl}

\and

\IEEEauthorblockN{Damien Anderson}
\IEEEauthorblockA{\textit{Computer and Information}\\ \textit{Sciences Department} \\
\textit{University of Strathclyde}\\
Glasgow, UK \\
damien.anderson@strath.ac.uk}

\and

\IEEEauthorblockN{Ahmed Khalifa}
\IEEEauthorblockA{\textit{Game Innovation Lab}\\
\textit{Tandon School of Engineering}\\
\textit{New York University}\\
New York, USA \\
ahmed.khalifa@nyu.edu}

\and

\IEEEauthorblockN{John Levine}
\IEEEauthorblockA{\textit{Computer and Information}\\ \textit{Science Department}\\
\textit{University of Strathclyde}\\
Glasgow, UK\\
john.levine@strath.ac.uk}

\and

\IEEEauthorblockN{Jochen Renz}
\IEEEauthorblockA{\textit{Research School of}\\
\textit{Computer Science}\\
\textit{Australian National University}\\
Canberra, Australia \\
jochen.renz@anu.edu.au}

\and

\IEEEauthorblockN{Julian Togelius}
\IEEEauthorblockA{\textit{Game Innovation Lab} \\
\textit{Tandon School of Engineering}\\
\textit{New York University}\\
New York, USA \\
julian@togelius.com}

\and

\IEEEauthorblockN{Christoph Salge}
\IEEEauthorblockA{\textit{Game Innovation Lab} \\
\textit{Tandon School of Engineering}\\
\textit{New York University}\\
New York, USA \\
c.salge@herts.ac.uk}

}

\maketitle

\begin{abstract}
This paper introduces an information-theoretic method for selecting a subset of problems which gives the most information about a group of problem-solving algorithms. This method was tested on the games in the General Video Game AI (GVGAI) framework, allowing us to identify a smaller set of games that still gives a large amount of information about the abilities of different game-playing agents. This approach can be used to make agent testing more efficient. We can achieve almost as good discriminatory accuracy when testing on only a handful of games as when testing on more than a hundred games, something which is often computationally infeasible. Furthermore, this method can be extended to study the dimensions of the effective variance in game design between these games, allowing us to identify which games differentiate between agents in the most complementary ways.
\end{abstract}


\section{Introduction}
Competitions and challenges are regularly used within AI as a way of evaluating algorithms, and also for promoting interest into specific problems. One design strategy to keep algorithms from over-specializing is to have competitions that are an ensemble of several different games or problems. Examples of challenges where this is the case include the GVGAI~\cite{perez20162014}, ALE~\cite{bellemare2013arcade} and many of the numerous Kaggle competitions~\cite{carpenter2011may}, each of which have hundreds of separate problems. There are also other sets of machine learning benchmarks that contain a multitude of disparate tasks, such as the OpenAI Gym~\cite{openaigym} or the UCI repository of supervised learning tasks~\cite{uclrep}.

However, this collection-of-problems approach has its own challenges. For large sets of possible problems it can be impractical, expensive or even impossible to evaluate a new algorithm on every instance within this set. Comparing a new algorithm with the state of the art on the full set of problems can require immense computational resources, which are not available to many researchers. So, what is a good subset of an existing collection that preserves the discriminatory power of the original test set? Another consideration is that several of the mentioned competitions like to include new problems every round to keep things interesting. But do the added problems really provide new and interesting challenges, or are they redundant compared to the existing set of tasks? Can we develop a formal way to test if a new problem adds something to the current field of the competition?

In this paper we will try to address these questions in general - and provide an application to the General Video Game AI (GVGAI) framework as an example and test of our method. The GVGAI library includes more than 100 mini video games~\cite{bontrager2016matching}, and several dozen agents that can play these games~\cite{soemers2016enhancements,gaina2017rolling,weinstein2012bandit,perez2016analyzing,mendes2016hyper} have been submitted to the associated GVGAI competition~\cite{perez2016general}.

Looking at the actual playing performance for a range of AIs and games allows us to make an important point: While the GVGAI competition produces a winner every year, it is not trivial to determine which of the competitors is the best algorithm in general. As we shall demonstrate later in this paper, there are games which produce a score distribution that is heavily anti-correlated to the majority of games in the set. A simple singular value decomposition \cite{githubPage} allows us to select a set of games that makes nearly every participating algorithm win, i.e. the selection of games can control which algorithms perform best. Furthermore, we need to keep in mind that the circa 100 games in the GVGAI game repository are only a subset of all possible games, so who is to say that they do not offer an extremely biased subset that favours a particular approach to AI? This problem is not easily solvable, and we bring it up mainly to demonstrate that asking for a subset that helps us identify the best algorithm is somewhat ill posed, if we have no guarantees that the larger set is at least a representative sample of all possible games. What we can do, though, is to find the subset that is best at differentiating among the set of existing algorithms. This selects a set of interesting problems, i.e. those that are best at producing different results for the current state of the art, and as such should also be of interest to those testing variations of existing AIs, wanting a quick check in comparison to existing problems.

To this end we propose an information-theoretic measure for determining which problems are best at telling a given set of algorithms apart; a measure that also takes into account the concept of noise when analyzing performance measures. By recursively applying this measure, we can find problems that are maximally informative considering previously selected problems, meaning that we can identify problems that discriminate among a set of algorithms in different ways \cite{martinez2016ai}. We should point out that our proposed measure is different in both motivation and procedure to those used in item response theory (IRT) \cite{irt1,irt2}, as IRT aims to estimate how good a given problem is at measuring the performance of an algorithm, while our approach seeks to identify problems that discriminate the most between different algorithms.

This measure should also be able to assist with the design of new problems for existing challenge collections. If a new problem adds little in terms of discrimination for an active field of algorithms, it remains doubtful that it impacts the challenge in any meaningful way. In contrast, a good addition should be placed highly in our ranking of discriminatory problems, or at least provide some new information about the existing set of algorithms.

The rest of the paper is structured as follows. We first introduce the GVGAI competition, and the methods used for data collection. We then provide a simple correlation analysis for both score and win rates to illustrate the structure of this, and potentially other, problem sets. This should convince the reader that a discriminatory subset is the best we can do. We then introduce the Information Gain Analysis method for finding such a subset - and argue that this is one principled and unbiased way to do this - given the stated assumptions. We also discuss how this method extends to arbitrary measures beyond score and win rate. Finally, we apply said method to the GVGAI dataset, and determine the top 10 most discriminatory problems, for a given set of games and algorithms. We then discuss the implications and limitations of this approach. Here it is important to keep in mind, that this measure addresses a specific question - and that it has specific features that we list. Other measures exist, or could be designed, that have a different set of features, and it is very much up to the aims of each particular researcher which measures are relevant to them.

\section{Background}
The GVGAI competition has been running annually since 2014 and provides a Video Game Description Language (VGDL) with which to quickly design games~\cite{Ebner2013}, and a common API for agents to access those games~\cite{Levine2013}. Each year ten games are selected to evaluate the submitted agents, which often covers a wide range of game types from role-playing to puzzle games~\cite{perez20162014}. One of the key elements of this competition is that the games being played by the agents for each year's competition are unknown to both the developers and agents beforehand. Many of the GVGAI games and agents include some form of stochasticity, meaning that performance evaluation is inherently noisy. Playing a GVGAI game also gives two signals of performance, whether an agent won the game or not and what score was obtained. The competition currently offers multiple tracks, including a planning track~\cite{perez20162014}, which provides a forward model for analyzing future game states, and a learning track which removes the forward model but allocates a training time to agents before submission~\cite{learningtrack}. For this paper, we only consider the games and agents used in the single-player planning track.

One of the main observations when evaluating GVGAI agents across multiple games is that they typically produce very different performance distributions \cite{bontrager2016matching}. It is also the case that there exist some games where nearly all agents either win or lose, in which case the score each agent achieved for the game would be the deciding value. Conceptually, it seems that games which offer a large spread of performance values would be best at discriminating between good and bad agents in a competition, but it was also clear from this prior analysis that not all games are won by the same agents. While a few previous papers have investigated the performance of agents on certain GVGAI games~\cite{Nelson2016,bontrager2016matching}, none have investigated the different discrimination profiles presented by the full game corpus, or how this information could be used to help design better agents and games in the future. The fact that different GVGAI games pose different types of problems to agents, may lead to biases towards a particular type of algorithm when selecting game subsets. Understanding what bias may exist in a given set of games, and being able to select ten games which minimize any particular bias, is desirable for ensuring that the GVGAI competition is genuinely evaluating general problem-solving capabilities.

\section{Data Collection}
The first step towards analyzing different GVGAI games is to collect data from various playthroughs using a collection of agents. We used 27 commonly available agents, which were some of the top performing entries in the previous GVGAI competitions over the last five years. Table 1 provides the names of these agents. The games that were used consist of the full corpus of 102 GVGAI games that are currently available (at the time of writing), plus an additional six deceptive GVGAI games introduced by Anderson et al.~\cite{Anderson2018}, giving a total of 108 games.

Similar to the GVGAI competition format, each agent has 40 milliseconds to perform each action and runs for at most 2000 time steps. To replicate the competition environment, we ran the agents using 243 CPU cores with 2.6 GHz and 8 GB of memory. Each successful playthrough of a game that resulted in either a win or loss without any crashes produces one unit of data, containing the information $\lbrack agent, game, score, win/lose\rbrack$. 
Unfortunately, some of the agents occasionally crashed on certain games due to changes that the GVGAI framework has received over the years, so not all the agents have the same amount of the generated data. A total of 3,990,760 successful playthroughs were recorded across all agents and games, with an average of 1,368.6 data samples per game-agent pair. The number of data samples obtained for each agent, averaged across all games, is shown in Table 1.

\begin{table}[htb]
\centering
\begin{tabular}{| l | r |}
\hline
Agent Name & Average Data per Game\\
\hline
\hline
jaydee & 1,202.7\\
MH2015 & 4,184.9\\
mrtndwrd & 1,029.8\\
CatLinux & 1,588.8\\
ICELab & 857.8\\
thorbjrn & 1,358.4\\
TomVodo & 1,030\\
sampleRHEA & 1,447.9\\
NovelTS & 1,307.3\\
evolutionStrategies & 1,390.7\\
simulatedAnnealing & 1,361.8\\
adrienctx & 1,117.6\\
sampleMCTS & 1,184.1\\
SJA862 & 2,416.2\\
greedySearch & 1,421.4\\
AtheneAI & 731\\
muzzle & 1,414.5\\
YBCriber & 1,037.6\\
NovTea & 1,296.2\\
MaastCTS2 & 1,364.3\\
Return42 & 1,261.7\\
aStar & 1,226.2\\
SJA86 & 1,130.8\\
bladerunner & 994.1\\
TeamTopbug & 1,243.7\\
Number27 & 1,183.1\\
hillClimber & 1,514.1\\
\hline
\hline
Minimum & 731\\
\hline
Average & 1,368.6\\
\hline
Maximum & 4,184.9\\
 \hline
\end{tabular}

\caption{The average number of data samples (playthroughs) per game obtained for each agent during data collection.}
\label{tab:playoutsPerGame}

\end{table}

\begin{figure*}
    \centering
    \begin{subfigure}[t]{.58\textwidth}
  	    \centering
	    \includegraphics[width=\linewidth]{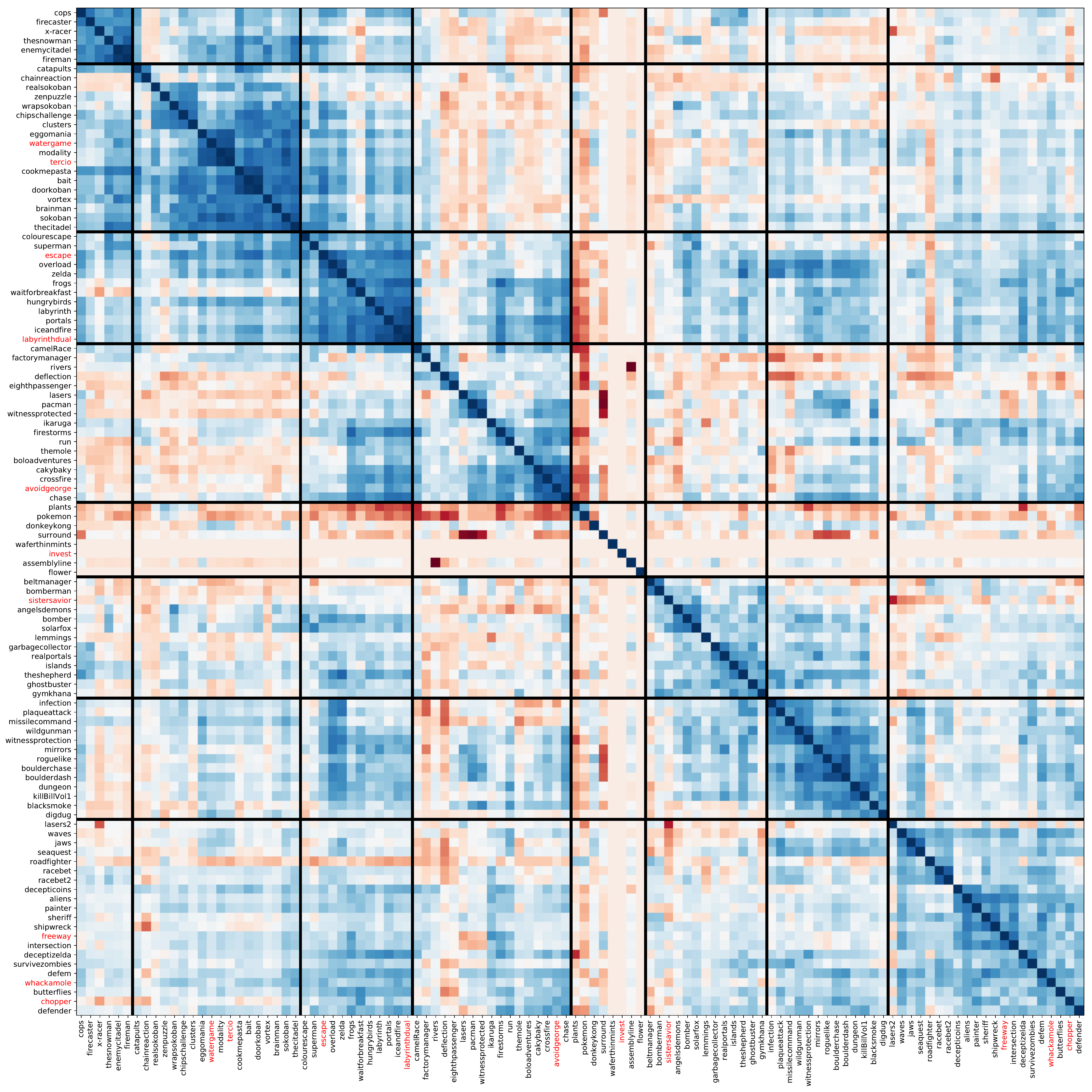}
        \caption{Win-rate correlation matrix.}
  	    \label{fig:correlationWin}
    \end{subfigure}
    \begin{subfigure}[t]{.58\textwidth}
  	    \centering
	    \includegraphics[width=\linewidth]{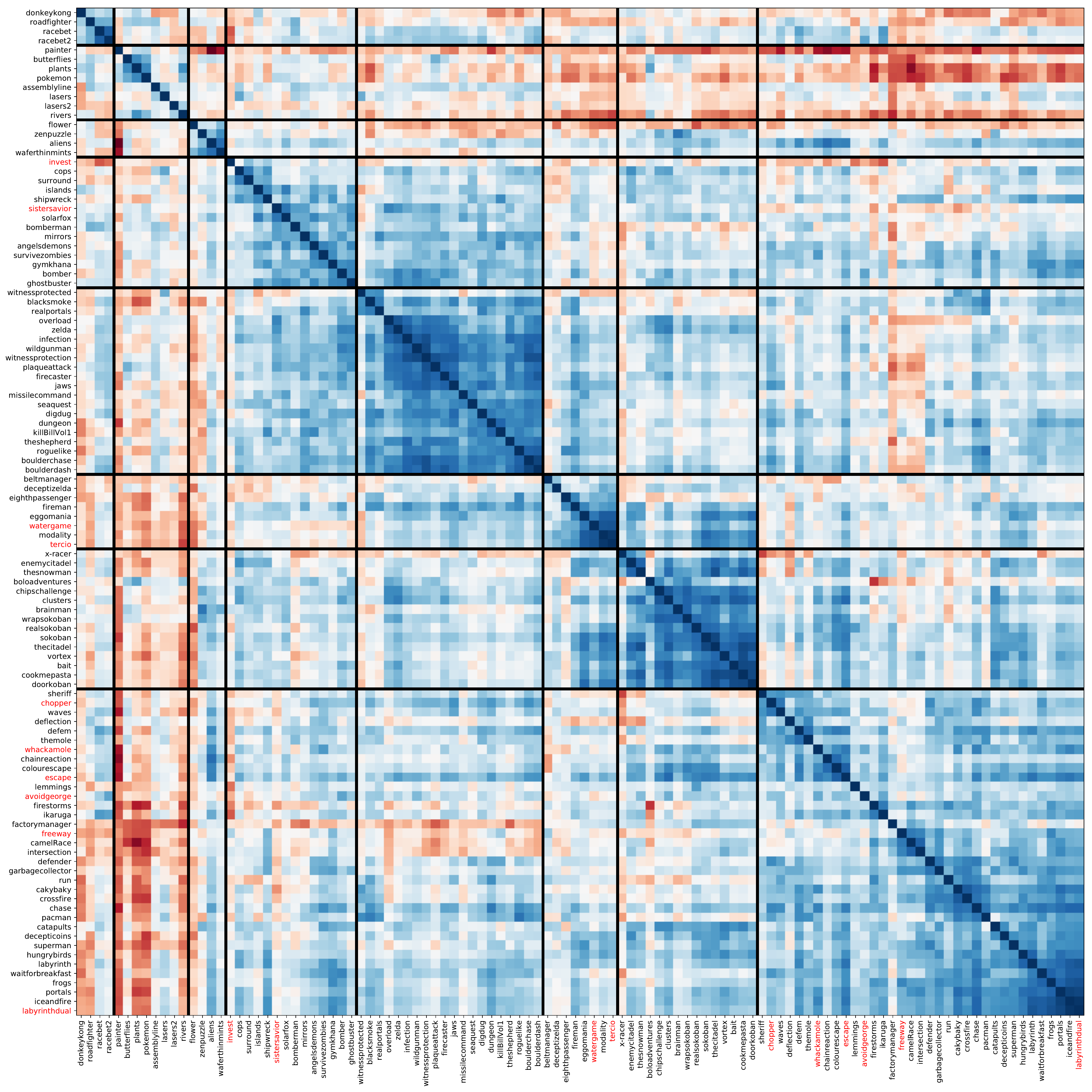}
        \caption{Score correlation matrix.}
  	    \label{fig:correlationScore}
    \end{subfigure}
    \caption{Correlation matrices between every game in the framework. Figure~\ref{fig:correlationWin} is based on the agents' win-rates, while Figure~\ref{fig:correlationScore} is based on the agents' scores. The games are sorted based on the result of a hierarchical clusterging algorithm.}
    \label{fig:correlation}
\end{figure*}

\section{Correlation Analysis}
Before describing our information-theoretic method we first present a preliminary analysis into the correlations between games in terms of agent performance, using both win-rate and score as performance measures. This helps to contextualize the problem space more and to visualize that different games measure different things, as games which have similar performance patterns should have similar problem characteristics. The resulting correlation matrices are then used for clustering, and these clusters are analyzed for meaningful similarities between games. While this approach does have several limitations that make it impractical for selecting a subset of highly discriminatory problems, this initial analysis has its own benefits and applications beyond simply providing a better domain understanding.

In the video game industry, similar video games are usually grouped under a specific category that is defined by common gameplay characteristics, referred to as a game genre. Games in the GVGAI framework are mostly ports of known video games, meaning that we can often find genre relations between them. However, attempting to group games by their genres does not necessarily indicate that similar problem-solving capabilities are required to solve them. We, therefore, took a more formal and robust approach for identifying correlations between games based on agent performance.

For this analysis, we calculated the correlation matrix between all 108 games in our sample using either the agents' win-rates or scores. Figure~\ref{fig:correlation} shows these correlation matrices where blue means high correlation, red means high anti-correlation, and white means no correlation. To simplify the task of analyzing such a large matrix, we clustered their values using an agglomerative hierarchical clustering algorithm - using the inverse of covariance matrix values as distance and a threshold 0.8 - and selected clusters that minimize the variance between the games within each cluster. The different clusters are represented by the black vertical or horizontal lines and are ordered (and subsequently referred to) in terms of their location from the left/top of the matrix.

Figure~\ref{fig:correlationWin} shows the correlation matrix using the agents' win-rates. Using this, we can see that games in the fifth cluster have a low anti-correlation to the rest of the games in the framework. These games are characterized by either being very hard to beat (plants) or not having a winning condition (invest). By analyzing the clusters row by row, we can see that the win-rates of most games are not highly correlated except for the first three clusters. Most of the games within these first three clusters appear to be puzzle games (zenpuzzle, sokoban, etc.). These types of games are typically characterized by the need for long-term planning to solve them, which likely causes their win-rates to be highly similar.

Figure~\ref{fig:correlationScore} shows the correlation matrix using the agents' scores. Using this, we can see that the score distributions between most of the games are similar (the matrix is mostly blue). This was not surprising as we already know that most of the games in the framework are designed to have a score distribution that reflects the progress of the agents in the game (good states have high scores, while bad states have low scores). The only exception to this is the first three clusters, which are highly anti-correlated with every game in the framework that isn't within its own cluster. These games appear to be characterized by a delayed score distribution (score only received near the end of the game) which makes them very different from the other games that provide rewards for incremental steps closer to the solution.

While our presented correlation matrices could be used to roughly identify a collection of games with decent discriminatory performance by selecting a game from each cluster, this approach has several limitations. Not only is it difficult to tell which games in each cluster would provide the most information, but neither the fact that certain agent's performance on the same game can vary dramatically between attempts, nor that two distinct performance measures are available, are taken into account. However, these correlation results can certainly be useful in other areas, such as for allowing game designers to detect which games are similar in terms of agent performance. Identifying games that present unique performance distributions could help in designing additional games that fit entirely new or underrepresented clusters. Accomplishing this would increase the overall discrimination potential of our total game set, and thus also increase the total amount of information that could be achieved from a subset of games (i.e. allows our information-theoretic measure proposed in the following section to be even more effective).

\section{Information Gain Analysis}

In this section, we analyze the information provided by each of our 108 sample games. Information here is used in the sense of Shannon Information Theory~\cite{eoit}, and the information gain of a game is the average reduction in uncertainty regarding what algorithm we are testing, given the score and/or win-rate performance of that algorithm. This information gain measure can then be used to identify a benchmark set of games that provides us with the maximum information about our agents.

While it is possible to compute information gain on discretized data by first binning the mean performances of the different agents, this is problematic for two reasons. First, as long as all the agents' results are at least somewhat different, it would be theoretically possible to obtain all information from just a single game. This situation would make calculating the information gain highly redundant, as nearly all of the games would give us the same, maximal amount of information. Second, this approach disregards any noise within the measuring process. As an example, if we assume that the average results for two agents are .49 and .50 when playing a specific game, then a discretized information gain analysis would give these as two separate outcomes (assuming we binned to the nearest .01 value). This approach does not take into account the fact that repeated measurements would likely produce slightly different results, varied by some noise. Consequently, a game that gives us average results for two agents of .1 and .9, rather than our previous example of .49 and .50, would be much better suited to tell two agents apart, as the scores are likely to be significantly different even when taking noise into account. In essence, games with agent results that are furthest apart and with the lowest noise, provide the most information.

The following information gain formalism is an attempt to accurately measure this difference by modelling the noise within the agents' performances as a Gaussian distribution. This approach calculates the information gain for a specific game $g$. Let us first define a few terms:

\begin{itemize}
\item $\mathcal{A}$: The set of all algorithms, $a$
\item $a_n$: A specific algorithm, having an average performance of $\mu_n$, with a variance of $\sigma_n$, for the game in question.
\end{itemize}

 We assume that $A$ is equally distributed, $p(a) = 1/{|A|}$, so everything else being equal we have a similar chance to encounter any of the algorithms in $A$, and our a priori assumption about $A$ reflects this. This allows us to approximate the conditional probability $p(a_2|a_1)$. This probability expresses how likely it is that we are observing the result of algorithm $a_2$, if we are in fact observing the average performance for algorithm $a_1$. In other words, how well does $a_2$ work as an explanation for what we see from $a_1$. 

Equation 1 approximates this probability. It assumes that observations of specific performances are normally distributed, parameterised by the means and variances from their actual results. The upper part of the equation is the probability density function for a normal distribution based on $a_2$, computing how likely a result equal to the mean of $a_1$ is. The denominator is a normalization sum over all possible algorithms, ensuring that the overall probabilities sum to one. 
\begin{equation}
p(a_2|a_1) \approx \frac{\frac{\exp{\Big(-\frac{(\mu_1-\mu_2)^2}{2{(\sigma_2+\sigma_1)}^2}}\Big)}{\sqrt[]{2\pi {(\sigma_2+\sigma_1)}^2}}}{\sum\limits_{a \in \mathcal{A}}{\Bigg( \frac{\exp{\big(-\frac{(\mu_1-\mu_a)^2}{2{(\sigma_a+\sigma_1)}^2}}\big)}{\sqrt[]{2\pi{(\sigma_a+\sigma_1)}^2}}\Bigg)}}
\end{equation}

Computing the probabilities for all pairwise combinations of algorithms allows us to define a confusion matrix $C$ between $n$ different algorithms as:
\begin{equation}
C = 
 \begin{pmatrix}
  p(a_1|a_1) & p(a_2|a_1) & \cdots & p(a_n|a_1) \\
  p(a_1|a_2) & p(a_2|a_2) & \cdots & p(a_n|a_2) \\
  \vdots  & \vdots  & \ddots & \vdots  \\
  p(a_1|a_n) & p(a_2|a_n) & \cdots & p(a_n|a_n)
 \end{pmatrix}
\end{equation}

Each row of the matrix sums to 1, and each entry in the first row indicates our best guess for the actual algorithm given that we observed the mean of algorithm $a_1$. The matrix of conditional probabilities can then be seen as an error matrix for a channel defined by using the game in question as a measurement device. This allows us to compute the mutual information for this channel,assuming that the a priori distribution for $A$ is an equal distribution (as stated earlier). This value is equivalent to the amount of information we get about what algorithm is used from observing the average performance result. Formally, we can define this as the mutual information between your belief distribution $\hat{A}$  and the distribution $A$ of the actual algorithm $a \in \mathcal{A}$, expressed in Equation 3.
\begin{equation}
\begin{split}
I(\hat{A};A) = H(\hat{A}) - H(\hat{A}|A)\quad\quad\quad\quad\quad\quad\quad\quad\\
= log_2(|\mathcal{\hat{A}}|) - \sum_{a \in \mathcal{A}}{p(a)}\sum_{\hat{a} \in \mathcal{A}}{-p(\hat{a}|a)\log_2{{p(\hat{a}|a)}}}
\end{split}
\end{equation}

The a priori distribution of our beliefs $\hat{A}$, is an equal distribution. If we observe the average performance of the algorithm $a$ we get a distribution of $\hat{A}|a$, as defined by the confusion matrix. The average information gain of observing these results is the average difference in the entropy before observation $H(\hat{A})$ and after observation, $H(\hat{A}|A)$. The equal distribution reduces to the $\log$ of the states, so we only need to compute the conditional entropy. A higher value here is more desirable, as the best games should provide us with the most information.

Note also, that this algorithm is not designed to actually determine the algorithm in question, but to give us one scalar value that expresses how \textit{good}, on average, a specific game would be at determining this.

\subsection{Information gain for multiple games}

The previous formalism allows us to quantify how much information a single specific game can provide us with about what algorithm is being used, but the information gained from looking at two games is always less than or equal to the sum of the information gain from both games individually. This is because the performance in both games could provide us with similar information. Instead of using a naive Bayesian approach, which would count these contributions twice, we can directly compute a confusion matrix for a pair or any larger set of games $g \in G$ by extending the definition of the conditional probability to that presented in Equation \ref{eq:multiGames}.

\begin{equation}\label{eq:multiGames}
p(a_2|a_1) \approx \frac{\frac{\exp{\Big(-\sum\limits_{g \in G}\big(\frac{(\mu_{1,g}-\mu_{2,g})^2}{2(\sigma_{2,g}+\sigma_{1,g})^2}}\big)\Big)}{\prod\limits_{g \in G}\Big(\sqrt[]{2\pi (\sigma_{2,g}+\sigma_{1,g})^2}\Big)}}{\sum\limits_{a \in \mathcal{A}}\Bigg({\frac{\exp{\Big(-\sum\limits_{g \in G}\big(\frac{(\mu_{1,g}-\mu_{a,g})^2}{2(\sigma_{a,g}+\sigma_{1,g})^2}}\big)\Big)}{\prod\limits_{g \in G}\Big(\sqrt[]{2\pi (\sigma_{a,g}+\sigma_{1,g})^2}\Big)}}\Bigg)}
\end{equation}

This does rely on the assumption though, that the noise added to both performance measures is independently distributed. While the average performances of the algorithms might be correlated (which this measure accounts for), the independence of noise is a relatively safe assumption. The n-th performance measure of an algorithm in one game is unlikely to affect the n-th performance measure in another game. This is an issue when combining the measure for score and win-rate, as it is likely that these two measurements are correlated. A more faithful, but also more complex, approximation could be achieved by using the Mahalanobis distance \cite{mahalanobis1936generalized} instead of the sum of variances - but this has not been realized for our data here.

Using this new conditional probability definition allows us to compute the information gain for any subset of games, by just picking a suitable set $G$. The mutual information for the resulting confusion matrix is computed as usual. This means that the theoretical maximum information gain that any set of games could give is equal to $log_2(|A|)$.
In general, those games that offer different kinds  of information lose less information due to redundancy. 

\begin{table*}[!t]
\begin{center}

\begin{tabular}{| p{1.8cm} | p{1.55cm} | p{2.1cm} | p{1.55cm} | p{2.1cm} | p{1.55cm} || p{1.8cm} | p{2.2cm} |}
\hline
Game Name (win-rate) & Information gain & Game Name (score) & Information gain & Game Name (combined) & Information gain & Game Name (top 10) & Information gain (cumulative)\\ \hline
freeway & 1.17484168 & invest & 1.62405816 & freeway & 1.89430152 & freeway & 1.89430152\\ \hline
labyrinth & 1.10088062 & intersection & 1.13955416 & invest & 1.62405816 & invest & 3.08236771\\ \hline
tercio & 1.10018133 & freeway & 1.13619392 & intersection & 1.59362941 & labyrinthdual & 3.81992620\\ \hline
labyrinthdual & 1.08531707 & tercio & 1.10018133 & chopper & 1.48524965 & tercio & 4.22563462\\ \hline
iceandfire & 1.07275305 & watergame & 0.89206793 & tercio & 1.44693431 & sistersavior & 4.40856274\\ \hline
chopper & 1.06542656 & cops & 0.88658183 & labyrinthdual & 1.42090667 & avoidgeorge & 4.54036694\\ \hline
doorkoban & 0.98911214 & flower & 0.86746818 & iceandfire & 1.32455879 & escape & 4.60252506\\ \hline
hungrybirds & 0.91886839 & waitforbreakfast & 0.80128373 & hungrybirds & 1.32100004 & whackamole & 4.64444512\\ \hline
watergame & 0.89206793 & labyrinth & 0.78021437 & waitforbreakfast & 1.28983481 & chopper & 4.67138328\\ \hline
escape & 0.87721725 & realportals & 0.73246317 & doorkoban & 1.28593860 & watergame & 4.68457480\\ \hline
\end{tabular}
\end{center}
\caption{The games with the highest information gain (using win-rate, score or both combined as measure of performance), as well as the top 10 games which collectively provide the highest information gain.}
\label{tab:dataPerGame}

\end{table*}

\subsection{Combine win-rate and score together}

Using the previous equations, we can calculate the information gain for a particular game or set of games, using either the win-rate or score as the measure of performance. However, it is also possible to calculate the total information gain based on both win-rate and score combined. To do this, we treat each of these cases as a separate game (i.e., for a particular game $g_{i}$ there are two variants, one where the win-rate is used as the measure of performance $g_{i,w}$ and one where the score is used $g_{i,s}$). Since the distance is scaled by the variance, both win-rate and score can be translated in the same way as information. We can then use Equation 4 to calculate the total combined information gain of the game $g_{i}$ by setting $G = [g_{i,w} , g_{i,s}]$. This means we can create a single confusion matrix for each game that encompasses both the win-rates and scores of all agents. The first six columns of Table~\ref{tab:dataPerGame} show the 10 games with the highest information gain when using either the win-rate, score, or both of these combined as the measure of performance. In general, this approach allows for the combination of any scalar values expressed by the game, and can, therefore, be applied to a range of different gaming benchmarks, even those where games have entirely different performance measures. 


\subsection{Top ten games}

By initially selecting the game that provides the largest information gain (based on both win-rate and score combined) and then recursively selecting the game that adds the most information to the already selected games, we can create a set of 10 games that provide the most information possible. The rightmost two columns of Table~\ref{tab:dataPerGame} provide the 10 games that were chosen for this set in the order they were selected, along with the total cumulative information gain of the set after each game was added. These 10 games are also highlighted in red in Figure~\ref{fig:correlation}, and we can see that the selected games mostly come from different clusters.

The theoretical maximum information gain that any set of games could give is roughly 4.75 = $log_{2}(27)$, so we can see from these results that after selecting only 3 or 4 games we can already get the majority of information about which agent is playing.
It is worth noting that this set of games is not simply the ten games that individually provide the most information, as some of these games likely provide the same ``kind'' of information. For example, the game intersection had the third highest information gain when looking at each game individually but was not selected for our top 10 games set. This is likely because it provides the same information as one of the previously selected games. By looking at how this game is played and our correlation matrices in Figure~\ref{fig:correlation}, it would appear that this game is very close to that of the game freeway and would likely give similar information.

A similar effect also appears when we look at both score and win-rate for the same game, as these two values are often, but not always, correlated. We can compare the information gain provided by just the win-rate or score for certain games, versus the combined information gain from using both. When looking at each of these performance measures separately it appears that Invest has the highest information gain, which is likely due to the large variation in possible scores that agents could achieve. However, as agents will always lose this game, either by spending too much money or the time limit expiring, the win-rate provides no information gain at all. Freeway, on the other hand, has a high information gain when using either win-rate or score, allowing it to have a combined information gain that is higher than Invest. It is worth reiterating that the combined information gain for a game is not simply the sum of its individual parameters, as some information may be shared between the different performance measures (calculation for combined information gain is sub-additive).

\section{Conclusions}
In this paper, we present an information-theoretic method for selecting which problems to evaluate a given algorithm on. We extend the notion of information gain to combine two feedback signals (win-rate and score) and to work on continuous values with modelled noise.
This method selects problems that are both discriminating by themselves, i.e. the performance of the different algorithms is evenly distributed, and also measure different agent properties compared to the other selected problems, i.e. the performances between problems are not highly correlated. This is beneficial for constructing a suitable set of discriminatory problems, as it allows us to identify if we already have several problems that provide similar challenges. This is particularly useful for the many cases where it is computationally infeasible to test a new algorithm on all benchmark problems. Our method is generally applicable to any situation where algorithms need to be tested on many problems and is especially useful when the problems are noisy and/or have multiple performance metrics.

As part of developing this method, we analyzed the discriminatory capabilities of the games in the GVGAI framework, as well as the correlations between games in terms of agent performance. Our correlation analysis shows that there are substantial variations in agent performance between different GVGAI games, and the resulting correlation matrices can be used to cluster certain games together. Developing new GVGAI games that do not fit within these identified clusters would present an entirely new challenge for the current selection of agents, making them highly desirable. Games that have different discriminatory profiles from those that already exist would likely be more useful for investigating agent performance than those with similar profiles to previous games. In this way, our approach could also help to design new additions to existing challenges.
Finally, our approach can also provide us with a better understanding of the structure of a given problem set. For example, analyzing the set of GVGAI games reveals that several of them have ``deceptive'' qualities, where the score is not strongly correlated with the win-rate. 

We also demonstrate how this measure can be applied recursively to find a small set of games that gives us almost as much information about an agent as the full set of games would have. This will hopefully allow future developers and researchers to accurately compare the performance of their new agent against the current set of evaluated agents without the need for exhaustive testing on the full GVGAI corpus. New agents can be tested on a set of “exploratory experiments” to help gauge how well the agent may perform on more detailed experiments that consider the entire GVGAI game set. This will be especially important in the future as more and more games are added to the GVGAI game library, resulting in significantly increased benchmark evaluation times. The proposed approach can also be used to evaluate games that are selected for future GVGAI competitions, to ensure that they present a diverse range of problems.

\subsection{Considerations}
It is important to note that we do not propose that all researchers replace testing on complete benchmark sets with only using a subset; rather, we propose an approach for finding the right subset of benchmark problems when conducting initial studies. Such studies will, of course, need to be followed up by more extensive testing using larger benchmark sets. We also acknowledge that in some cases, other more tailored subsets will need to be used in order to test specific hypotheses.

We should also point out that the selected subset of games depends both on the original complete set of games and also the set of algorithms studied. Having a different set of agents could mean different games, that might previously have been too hard, would suddenly be more discriminatory. Similarly, adding additional games can affect which games provide us with redundant information. Because of this, while the specific games identified here are interesting today, they might very well change in the future. It is therefore meaningless to ask for the ``true'' set of most discriminatory games, as that will always be relative to the agents available.
However, we believe the more important contribution of this paper is the methodology we propose to select these games.

\subsection{Future Work}
While this method has been applied here on GVGAI games, it could also be used for other sets of problems and algorithms. Our approach can be generalized from just win-rates and scores to include any number of different outcome measures from other domains. This could include problems outside of the traditional game space, as long as the mean and variance of each algorithm's performance can be obtained. An obvious future application would be to analyze the performance of multiple deep reinforcement learning algorithms on the Atari games in the Arcade Learning Environment (ALE) framework and supervised learning algorithms tested on datasets associated with Kaggle competitions. Regarding our specific use case of applications for GVGAI, future work could involve expanding the evaluation criteria to include additional data from agent playthroughs, such as the time required to solve a level or the number of moves used, which may help us to better differentiate between agents.

Using the methods presented here to evaluate newly created games by how well and uniquely they distinguish among algorithms would also be worthwhile, yielding an iterative process where games under development can be tuned so as to increase their discriminative capacity. In the long run, this process could even be automated, with a stochastic search algorithm finding games that maximize unique discriminatory abilities. Previous work has created games optimized for differentiating between specific pairs of agents, and this procedure in combination with the method proposed here could conceivably work for differentiating among larger sets of agents~\cite{nielsen2015towards}. This would constitute significant progress towards automatically creating relevant benchmarking problems.

\section*{Acknowledgements}
Ahmed Khalifa acknowledges the financial support from NSF grant (Award number 1717324 - ``RI: Small: General Intelligence through Algorithm Invention and Selection.'').

\section*{Online Resources}
All data, results and code associated with this paper can be accessed online at:

https://github.com/stepmat/ContinuousInformationGain

\bibliographystyle{ieeetran}
\bibliography{sample-bibliography}

\end{document}